\RequirePackage[T1]{fontenc}
\documentclass[conference]{IEEEtran}
\usepackage{times}
\usepackage[numbers]{natbib}
\usepackage{graphicx}
\usepackage{capt-of}
\usepackage{caption}
\usepackage{booktabs}
\usepackage{array}
\usepackage{tabularx}
\usepackage{amsmath,amssymb}
\usepackage{xcolor}
\usepackage{microtype}
\usepackage{placeins}
\usepackage[hidelinks,breaklinks=true,bookmarks=false]{hyperref}

\IEEEoverridecommandlockouts                              

\title{\LARGE\bfseries
ForceDelta-VLA: Distilling Force-Conditioned Action\\Corrections for Contact-Rich Manipulation}

\author{
    Ju Dong$^{1}$,
    Yu Fu$^{1}$,
    Jian Chen$^{2}$,
    Yimeng Liu$^{3}$,
    Haocheng Zhao$^{3}$,
    Lei Zhang$^{1,\dagger}$,\\
    Kaixin Bai$^{1}$,
    Liding Zhang$^{3}$,
    Diwen Zheng$^{3}$,
    Alois Christian Knoll$^{3}$,
    Angela P. Schoellig$^{3}$,
    Jianwei Zhang$^{1}$
    \thanks{$^{\dagger}$Corresponding author: lei.zhang-1@studium.uni-hamburg.de.}
    \thanks{$^{1}$TAMS (Technical Aspects of Multimodal Systems), Department of Informatics, University of Hamburg, Hamburg, Germany.}
    \thanks{$^{2}$University of Science and Technology of China, Hefei, China.}
    \thanks{$^{3}$Technical University of Munich, Germany.}
}

\begin{document}
    \makeatletter
    \let\@oldmaketitle\@maketitle
    \renewcommand{\@maketitle}{\@oldmaketitle
      \centering
      \includegraphics[width=0.94\textwidth]{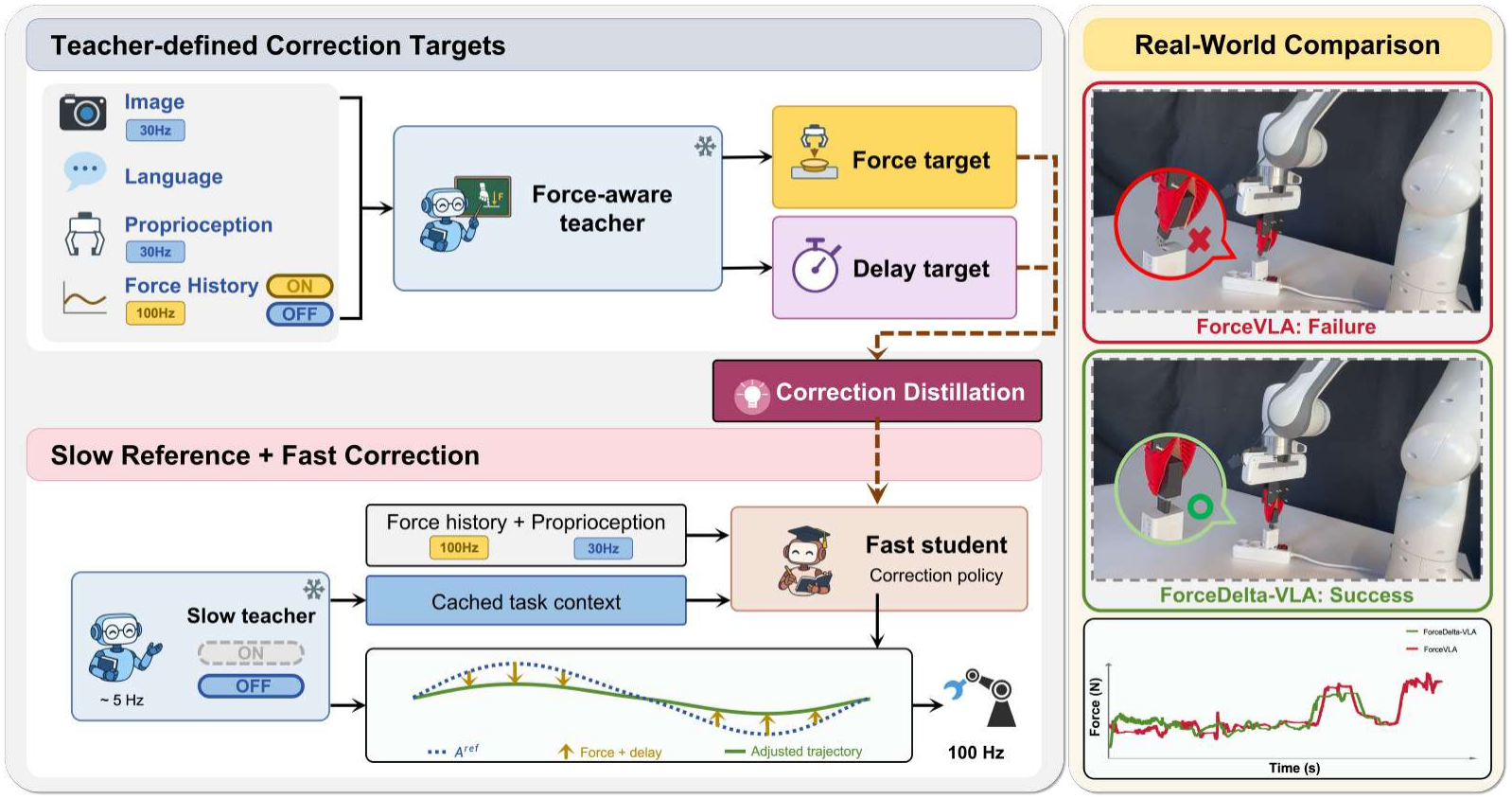}
      \captionof{figure}{ForceDelta-VLA overview. Left: A frozen teacher defines force and delay targets for correction distillation. The fast student uses recent force history and proprioception to adjust the slow teacher's reference actions while reusing its cached task context. Right: USB Insertion examples and corresponding contact-force traces for ForceVLA and ForceDelta-VLA.}
      \label{fig:teaser}
      \vspace{-0.1in}
      }
\makeatother
\maketitle

\thispagestyle{empty}
\pagestyle{empty}
\setcounter{figure}{1}
\begin{abstract}
Force-aware Vision-Language-Action (VLA) policies improve contact-rich manipulation, but typically combine task-level motion and contact-dependent adjustment in a single action prediction. Demonstrations provide no explicit labels for decomposing that prediction into a reusable reference action and a correction. We present ForceDelta-VLA, a correction-distillation framework that constructs an explicit force-correction target using paired predictions from a frozen teacher's force-conditioned and learned force-agnostic modes. A separate delay-correction target accounts for reference-action mismatch and the change in reference state. Training uses asynchronous schedule replay with the cached task context available during execution. The resulting lightweight policy adjusts the reference actions using recent force history and robot state, responding to contact changes between reference-action updates without regenerating complete action chunks. Across nine single-arm and bimanual contact-rich tasks, ForceDelta-VLA achieves an $82.2\%$ mean success rate, compared with $54.4\%$ for the original ForceVLA baseline. Direct execution of our Stage-1 Temporal Teacher achieves $70.6\%$. Relative to ForceVLA, the complete system reduces mean peak contact force over successful trials by approximately $26\%$ on both platforms.
\end{abstract}

\section{Introduction}
Contact-rich manipulation requires actions to adapt as contact develops. During USB insertion, for example, the approach direction may remain useful across several control cycles, while local alignment must respond to newly observed contact. Force-aware Vision-Language-Action (VLA) policies incorporate interaction feedback into action generation~\cite{yu2025forcevlaenhancingvlamodels,li2026forcevla2unleashinghybridforceposition,zhang2025tavlaelucidatingdesignspace}, but typically predict task motion and contact-dependent adjustment together. Separating a reference action predicted by a slow pathway from a rapidly updated correction would allow contact feedback to affect execution without waiting for a new complete action chunk (Fig.~\ref{fig:teaser}).

Learning this correction from demonstrations is not straightforward. Demonstrations record the final action, without specifying which part belongs to the reference motion and which part is a contact-dependent adjustment. Different reference actions and corrections can add up to the same demonstrated action. A correction pathway therefore needs a defined supervision target in addition to a fast architecture.

Existing asynchronous policies separate computation or sensor-update rates~\cite{li2026favlaforceadaptivefastslowvla,wang2026lateforceacceleratingvla,vanjani2026damvladecoupledasynchronousmultimodal}. Residual reinforcement learning and corrective imitation obtain correction signals from rewards or extra on-policy interaction~\cite{ankile2024imitationrefinementresidual,xu2025compliantresidualdaggerimproving}. We address the supervision problem using only the existing demonstrations, while retaining asynchronous execution.

We present ForceDelta-VLA, which learns force and delay corrections to a reusable reference action. Paired predictions from a frozen teacher's force-conditioned and learned force-agnostic modes provide force-correction supervision. A separate delay target accounts for the difference between the earlier reference action and a current force-agnostic prediction, including the change in reference state. The trained lightweight policy uses recent force history and robot state to update these corrections independently of reference-action generation.

We evaluate ForceDelta-VLA on five single-arm and four bimanual real-robot tasks involving initial contact, precise insertion, motion against friction, and motion constrained by the environment.

The contributions of this work are:
\begin{enumerate}
    \item We introduce \textbf{force-agnostic correction distillation}, which uses only the original dataset to construct an explicit force-correction target from matched predictions of a frozen teacher's force-conditioned and learned force-agnostic modes;
    \item We develop a \textbf{state-consistent dual-correction policy} that separately learns force and delay corrections in a common action coordinate system and updates them independently of reference-action generation while reusing task context. The delay target includes both reference-action mismatch and reference-state alignment; and
    \item Real-robot evaluation across nine single-arm and bimanual tasks shows $82.2\%$ mean success, versus $54.4\%$ for ForceVLA and $70.6\%$ for direct execution of the Stage-1 Temporal Teacher.
\end{enumerate}

\section{Related Work}
\label{sec:related}
\subsection{Force-Aware Manipulation Policies}
Force-aware policies incorporate interaction feedback into action prediction~\cite{yu2025forcevlaenhancingvlamodels,li2026forcevla2unleashinghybridforceposition,zhang2025tavlaelucidatingdesignspace,li2026fmvlaforcebasedmemoryvisionlanguageaction,zhang2026craftadaptingvlamodels,liu2025forcemimicforcecentricimitationlearning}. These methods typically predict task motion and contact adjustment together. ForceDelta-VLA separates a reusable reference action from corrections driven by recent wrench feedback. This also differs from FD-VLA, which distills a force representation for inference without measured wrench feedback~\cite{zhao2026fdvlaforcedistilledvisionlanguageactionmodel}.

\subsection{Multi-Rate and Structured Manipulation Policies}
Prior policies separate sensing, action generation, and correction across different update rates~\cite{lee2025manipforceforceguidedpolicylearning,chen2026implicitrdpendtoendvisualforcediffusion,wang2026phaforcephasescheduledvisualforcepolicy,fang2026forcepolicylearninghybrid,zhuo2026fardp}. FAVLA and DAM-VLA extend multi-rate processing to VLAs~\cite{li2026favlaforceadaptivefastslowvla,vanjani2026damvladecoupledasynchronousmultimodal}, while RTC addresses continuity between asynchronously generated action chunks~\cite{black2025realtimechunking}. Our focus is on updating teacher-defined force and delay corrections while reusing a reference action, independently of complete action generation.

\subsection{Residual Policy Learning}
Residual policies add learned corrections to a base controller~\cite{johannink2018residualreinforcementlearningrobot}. Prior methods learn these corrections from rewards or on-policy human corrections~\cite{ankile2024imitationrefinementresidual,xu2025compliantresidualdaggerimproving,yu2026omnitactunepolicyagnosticrealworldrl,wang2026lateforceacceleratingvla}, or learn additive corrections through imitation learning~\cite{wang2026phaforcephasescheduledvisualforcepolicy}. ForceDelta-VLA instead constructs explicit correction targets from paired teacher predictions using the original demonstration dataset.

\subsection{Policy Distillation}
Robot-policy distillation often accelerates full-action generation~\cite{wang2024onestepdiffusionpolicyfast,dong2026flowsteprealtimemultimodal}. ForceDelta-VLA distills corrections and retains the generative teacher for reference actions. Classifier-free guidance combines conditioned and unconditioned predictions~\cite{ho2022classifierfreediffusionguidance}; guidance distillation transfers this combined output to a student~\cite{meng2023distillationguideddiffusionmodels}. The learned force-agnostic mode distinguishes unavailable force input from measured zero force, following missing-modality learning~\cite{ramazanova2024exploringmissingmodalitymultimodal,nezakati2024mmprobustmultimodallearning}.

\section{Method}\label{sec:method}

\begin{figure*}[!t]
    \centering
    \includegraphics[width=0.98\textwidth]{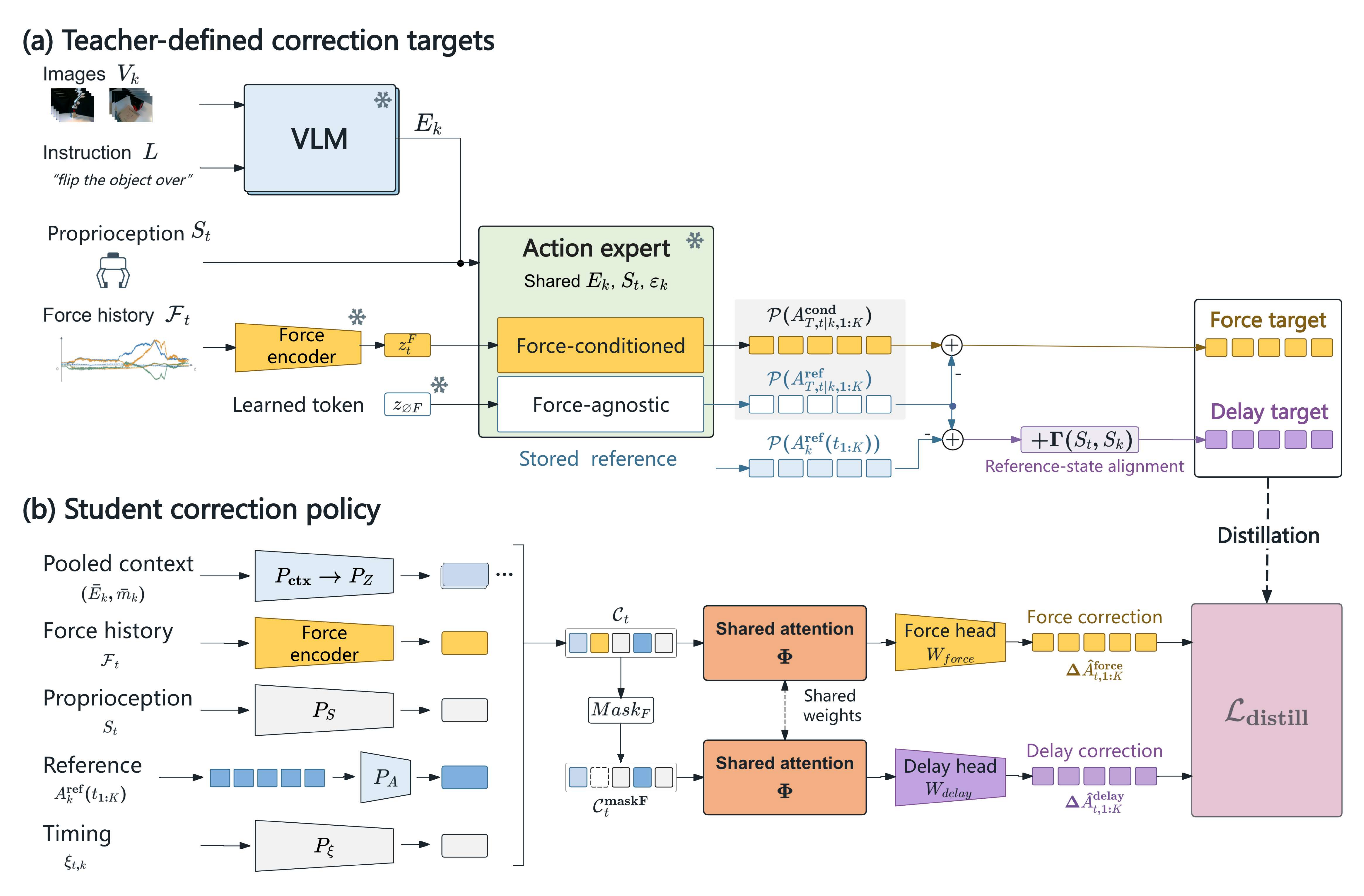}
    \caption{Correction distillation (training stage 3). (a) Paired predictions from the frozen teacher share context, proprioception, and sampling noise. The pose difference between the force-conditioned and force-agnostic predictions defines the force target. The delay target compares the current force-agnostic prediction with the stored reference and includes reference-state alignment. (b) Two evaluations of $\Phi$ share weights and the learned query $q_{\mathrm{corr}}$; only the delay evaluation masks the force token. Separate heads predict the two corrections, supervised by the teacher-defined targets.}
    \label{fig:network_structure}
\end{figure*}

\subsection{Overview and Notation}
\label{sec:problem}

ForceDelta-VLA separates reference-action generation from contact correction (Fig.~\ref{fig:network_structure}). The slow generative pathway predicts an action chunk without measured force input. We call this the reference action: a previously predicted motion that the fast pathway adjusts using force and delay corrections. Its action samples, visual--language context, and query state are stored for reuse. A lightweight pathway reads this cache together with the current robot state and recent wrench history, then predicts force and delay corrections. The executor combines these corrections with the reference action.

Training has three stages. We first train a temporal force-conditioned VLA teacher, then freeze it and learn a force-agnostic mode that supplies the reference action. With both modes fixed, paired predictions define correction targets, and a sampled asynchronous schedule provides the cached inputs used to train the correction policy.

We consider an offline dataset $\mathcal D$ of teleoperated manipulation trajectories. At time $t$, each trajectory provides visual observations $V_t$, a language instruction $L$, robot state $S_t$ (TCP pose), demonstrated action $A_t^*$, and the history of force/torque measurements up to time $t$
\begin{equation*}
    \mathcal F_t=\{(s,\mathbf w(s))\mid t-T_F\leq s\leq t\},
\end{equation*}
where $T_F$ is the history duration and $\mathbf w(s)=[\mathbf f(s);\boldsymbol\tau(s)]\in\mathbb R^6$ is the external end-effector force and torque (wrench) for each arm. Pose actions and wrench measurements use the corresponding robot-base frame. A force-conditioned teacher $T_\theta$ maps $(V_t,L,S_t,\mathcal F_t)$ to an $H$-step action chunk relative to $S_t$. Pose actions and corrections below use its normalized pose-action coordinates, with differences scaled by the action standard deviations.

Let $\mathcal P(\cdot)$ extract the pose component of an action. We decompose the commanded pose into a reference action and two learned corrections,
\begin{equation}
    \mathcal P(A^{\mathrm{cmd}})
    =\mathcal P(A^{\mathrm{ref}})
      +\Delta\hat A^{\mathrm{force}}
      +\Delta\hat A^{\mathrm{delay}}.
    \label{eq:action_comp}
\end{equation}
The force correction learns the paired teacher prediction difference. The delay correction compensates for the difference between previous and current reference predictions, including the difference between their reference states. Both terms modify only pose; the gripper command comes from the reference action.

Pose coordinates contain translation and a continuous 6D rotation representation~\cite{zhou2020continuityrotationrepresentationsneural}. Corrections are added in these coordinates, followed by Gram--Schmidt projection to a valid rotation. For bimanual tasks, we concatenate the inputs from both arms and jointly predict their pose actions. Representation and normalization details are provided in the supplementary material.

\subsection{Teacher and Force-Agnostic Reference Policy}
\label{sec:teacher_and_missing}

\subsubsection{Temporal force-conditioned teacher}

We build $T_\theta$ on ForceVLA~\cite{yu2025forcevlaenhancingvlamodels} and replace its instantaneous force embedding with a FAVLA-style causal temporal convolutional network (TCN)~\cite{li2026favlaforceadaptivefastslowvla} that encodes the wrench history,
\begin{equation}
    z_t^F=E_T^F(\mathcal F_t).
    \label{eq:teacher_tcn}
\end{equation}
The teacher is trained on $\mathcal D$ to predict
\begin{equation}
    A_{T,t,1:H}^{\mathrm{cond}}
    =T_\theta(V_t,L,S_t,z_t^F;\epsilon),
    \label{eq:conditioned_teacher}
\end{equation}
where $\epsilon$ is the initial flow noise. We freeze $\theta$ after training.

\subsubsection{Force-agnostic reference-action mode}

We distinguish an unavailable wrench history from a physically measured zero wrench. The force-agnostic mode handles unavailable force input: it replaces the force-history input with a learned token $z_{\varnothing F}$ and uses a rank-$\rho$ adapter $(W_{\mathrm{in}},W_{\mathrm{out}})$ to modify the action expert's pose output. We collect its trainable parameters as $\phi=\{z_{\varnothing F},W_{\mathrm{in}},W_{\mathrm{out}}\}$. The adapter modifies the pose channels of the action expert,
\begin{equation}
    v^p\leftarrow v^p+W_{\mathrm{out}}\,
    \sigma(W_{\mathrm{in}}h),
    \label{eq:missing_force_adapter}
\end{equation}
where $h$ is an action-expert hidden state, $v^p$ its pose-channel flow output, and $\sigma$ an element-wise activation function. The adapter is active only in the force-agnostic mode. With $\theta$ fixed, we train $\phi$ using the teacher's original flow-matching objective to learn the force-agnostic reference policy.
The reference-action prediction is
\begin{equation}
    A_{T,t,1:H}^{\mathrm{ref}}
    =T_{\theta,\phi}^{\mathrm{ref}}
      (V_t,L,S_t,z_{\varnothing F};\epsilon).
    \label{eq:teacher_missing}
\end{equation}

\subsection{Paired Correction Targets}
\label{sec:distillation}

\label{sec:teacher_differencing}

Target construction uses three predictions. The stored reference action was predicted at an earlier query state $S_k$ and is reused during execution. The current force-agnostic prediction is computed at $S_t$ without measured force input. The current force-conditioned prediction is computed at the same $S_t$ with recent force history. The latter two predictions are used only to construct training targets; neither replaces the stored reference during correction execution.

Let $t$ be a correction time and $k$ the latest reference-action query available at that time under the sampled schedule. Let $E_k$ be the visual--language prefix produced from $(V_k,L)$ when reference-action query $k$ starts at $t_k$ from state $S_k$ with flow noise $\epsilon_k$. Its force-agnostic action chunk defines the timestamped reference action, $A_k^{\mathrm{ref}}=A_{T,k}^{\mathrm{ref}}$. To match the information available to the deployed correction pathway, target generation reuses $E_k$ while updating the robot state and force history to time $t$:
\begin{align}
    A_{T,t|k,1:K}^{\mathrm{cond}}
    &=T_\theta(E_k,S_t,z_t^F;\epsilon_k)_{1:K},
    \label{eq:cached_conditioned_teacher}\\
    A_{T,t|k,1:K}^{\mathrm{ref}}
    &=T_{\theta,\phi}^{\mathrm{ref}}
      (E_k,S_t,z_{\varnothing F};\epsilon_k)_{1:K}.
    \label{eq:cached_missing_teacher}
\end{align}
Here $T(E_k,\cdot)$ reuses the cached prefix without encoding a current image. The reference action and the two current teacher queries share $E_k$ and $\epsilon_k$. For $j=1,\ldots,K\leq H$, the teacher-defined force-correction target is
\begin{equation}
    \Delta A_{T,t|k,j}^{\mathrm{force}}
    =\mathcal P\!\left(A_{T,t|k,j}^{\mathrm{cond}}\right)
     -\mathcal P\!\left(A_{T,t|k,j}^{\mathrm{ref}}\right).
    \label{eq:force_target}
\end{equation}
This target supervises the force-correction head.

The force target compares two predictions at the current state, while execution reuses a reference predicted earlier. The delay target accounts for this remaining difference and the change in reference state. For the $K$ correction samples at $t_j=t+(j-1)/f_A$, where $f_A$ is the action-sample rate within a chunk, the target is
\begin{equation}
    \Delta A_{T,k\rightarrow t,j}^{\mathrm{delay}}
    =\mathcal P\!\left(A_{T,t|k,j}^{\mathrm{ref}}\right)
      -\mathcal P\!\left(A_k^{\mathrm{ref}}(t_j)\right)
      +\Gamma(S_t,S_k).
    \label{eq:delay_target}
\end{equation}
Reference-state alignment expresses the current prediction relative to the state $S_k$ used to generate the stored reference. Here $A_k^{\mathrm{ref}}(t_j)$ is that reference interpolated at $t_j$, and $\Gamma(S_t,S_k)$ converts the state difference to normalized action coordinates. The delay head learns the entire target, including $\Gamma$.

\subsection{Correction Policy and Training}
\label{sec:residual_policy}

\subsubsection{Inputs and prediction}
Each reference-action query stores its action chunk $A_k^{\mathrm{ref}}$, query state $S_k$, and a pooled representation $\bar E_k$ of the visual--language prefix with its validity mask $\bar m_k$. A learned projector converts the pooled context into task-context tokens $Z_k^{\mathrm{ctx}}$. Timestamp-based interpolation provides the reference-action segment $A_k^{\mathrm{ref}}(t_{1:K})$ aligned with the correction outputs. A causal force encoder maps $\mathcal F_t$ to $u_t^F$, and timing features $\xi_{t,k}$ encode cache age and interpolation phase. The conditioning set is
\begin{align}
    \mathcal C_t=[&P_Z(Z_k^{\mathrm{ctx}});u_t^F;P_S(S_t);
    \nonumber\\[-2pt]
    &P_A(\operatorname{vec}(A_k^{\mathrm{ref}}(t_{1:K})));P_\xi(\xi_{t,k})].
    \label{eq:conditioning_set}
\end{align}
The projections map inputs to a shared token width, with learned type embeddings identifying each condition. $P_A$ maps the flattened $K$-step reference-action segment to one token. For the delay output, $\mathcal C_t^{\mathrm{maskF}}=\operatorname{Mask}_F(\mathcal C_t)$ masks the force token. Pooling, projection, and timing-feature details are provided in the supplementary material.

A shared attention module $\Phi$ combines the input tokens using a learned query $q_{\mathrm{corr}}$. We evaluate it twice with the same parameters: once with all conditions for the force correction, and once with the force token masked for the delay correction. Separate output heads predict the two $K$-step corrections,
\begin{align}
    \Delta\hat A^{\mathrm{force}}_{t,1:K}
    &=W_{\mathrm{force}}\Phi([\mathcal C_t;q_{\mathrm{corr}}])[q_{\mathrm{corr}}],
    \label{eq:force_output}\\
    \Delta\hat A^{\mathrm{delay}}_{t,1:K}
    &=W_{\mathrm{delay}}
      \Phi([\mathcal C_t^{\mathrm{maskF}};q_{\mathrm{corr}}])[q_{\mathrm{corr}}].
    \label{eq:delay_output}
\end{align}
The mask prevents direct dependence of the delay-correction output on the force token. The two outputs therefore come from separate forward passes through the shared module.

\subsubsection{Asynchronous schedule replay}
\label{sec:async_training}

We train the correction policy on reference actions delayed by inference. We sample query periods and inference latencies before target extraction and replay the same fixed schedule during policy optimization. At each sampled correction time, we select the most recently started reference query whose inference has completed. Its cached context, interpolated reference actions, and timing features are used to construct the student inputs and paired targets. Timestamp sampling and reference-selection details are provided in the supplementary material.

With normalized temporal weights $w_j\propto\beta^{j-1}$, $0<\beta\leq1$, and normalized-coordinate mean-squared loss $\ell_{\mathrm{pose}}$, the distillation objective is
\begin{align}
    \mathcal L_{\mathrm{distill}}=\sum_{j=1}^{K}w_j\big[&
      \ell_{\mathrm{pose}}(\Delta\hat A^{\mathrm{force}}_{t,j},
        \Delta A^{\mathrm{force}}_{T,t|k,j})
    \nonumber\\[-2pt]
      &+\lambda_{\mathrm{delay}}
        \ell_{\mathrm{pose}}(\Delta\hat A^{\mathrm{delay}}_{t,j},
        \Delta A^{\mathrm{delay}}_{T,k\rightarrow t,j})\big].
    \label{eq:correction_loss}
\end{align}
The coefficient $\lambda_{\mathrm{delay}}$ balances the force-correction and delay-correction losses.

\subsection{Asynchronous Deployment}
\label{sec:execution}

\begin{figure}[!htb]
    \centering
    \includegraphics[width=\columnwidth]{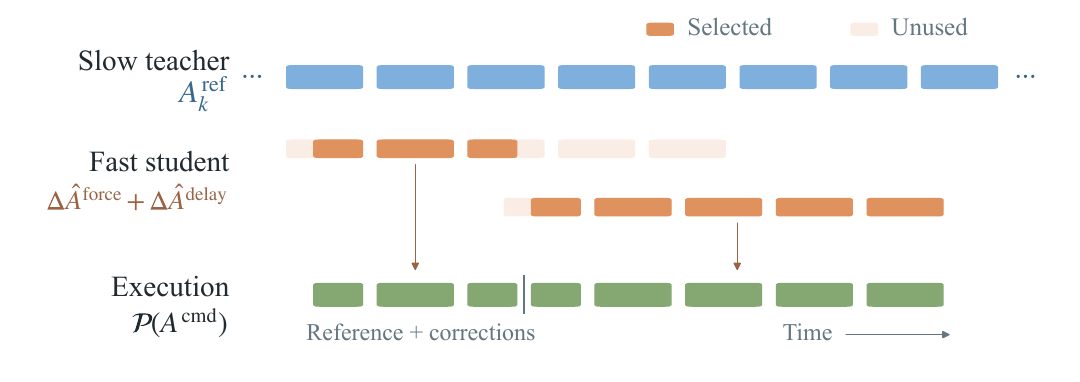}
    \caption{Deployment schematic. Two successive student queries reuse the same reference chunk. Selected correction steps are combined with the reference poses at the corresponding timestamps.}
    \label{fig:async_deployment}
\end{figure}

At deployment, the teacher and student run independently (Fig.~\ref{fig:async_deployment}). The teacher periodically refreshes the cached reference using its force-agnostic mode. Each student query reads the latest completed reference update and keeps that reference paired with its corrections throughout execution. Student queries continue while the teacher computes the next reference; queries started after that reference becomes available can adopt it.

At each control step, the executor selects the most recently started student query whose result is complete and still valid. For its active step $j$, it interpolates the reference at $t_j$, clips the two corrections separately, and composes the normalized pose using Eq.~\eqref{eq:action_comp}. The result is converted to an absolute command using the selected reference's query state $S_k$. Expired steps are skipped; a valid correction step is held until its next timestamp or a newer valid result becomes available. The timestamp-selection rules are given in the supplementary material.

\section{Experiments}
\label{sec:experiment}
\subsection{Experimental Setup}
\label{sec:exp_protocol}

\begin{figure}[!hb]
    \centering
    \begin{minipage}[c]{0.45\columnwidth}
        \centering
        \includegraphics[width=\linewidth]{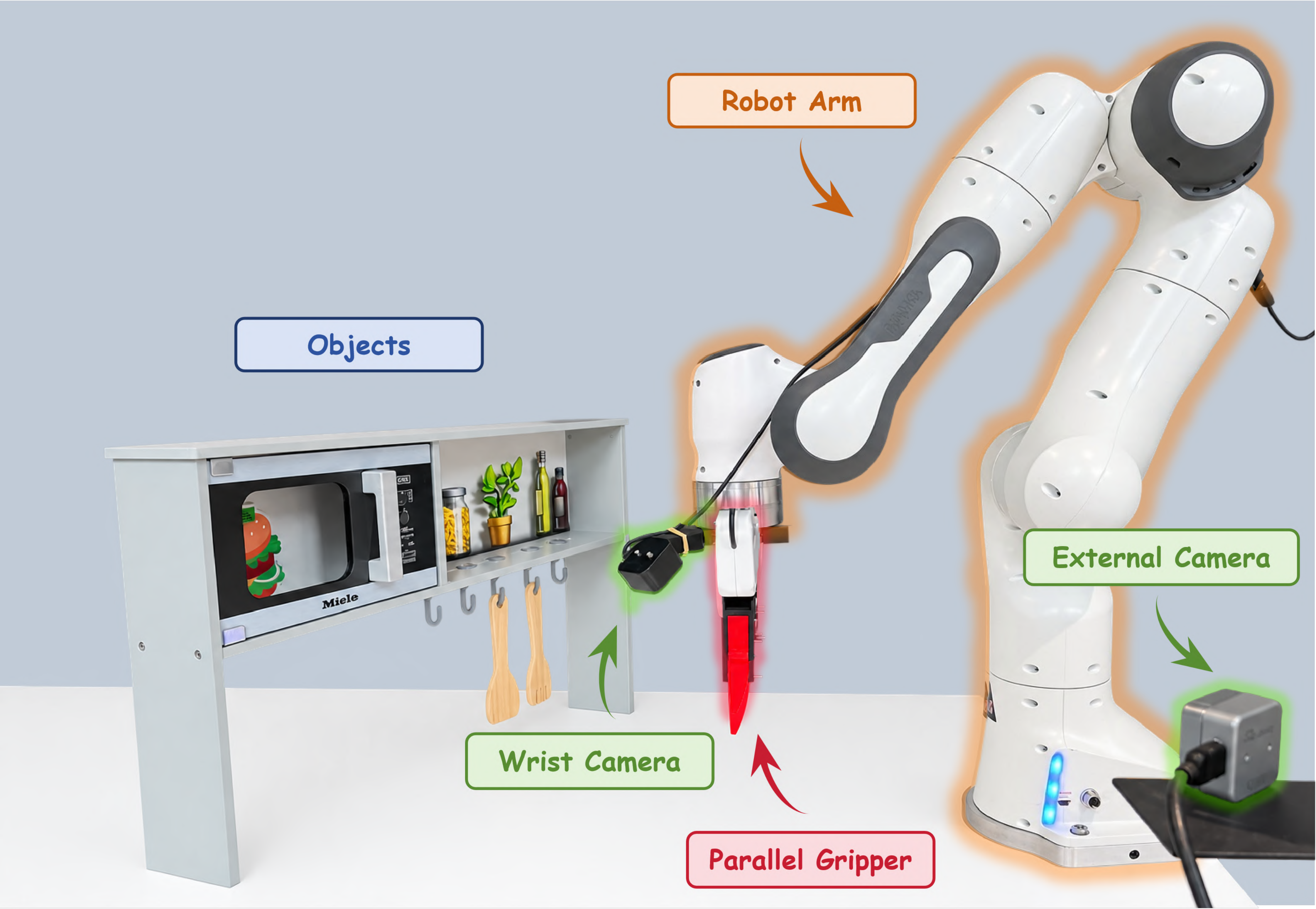}
    \end{minipage}\hspace{0.025\columnwidth}
    \begin{minipage}[c]{0.45\columnwidth}
        \centering
        \includegraphics[width=\linewidth]{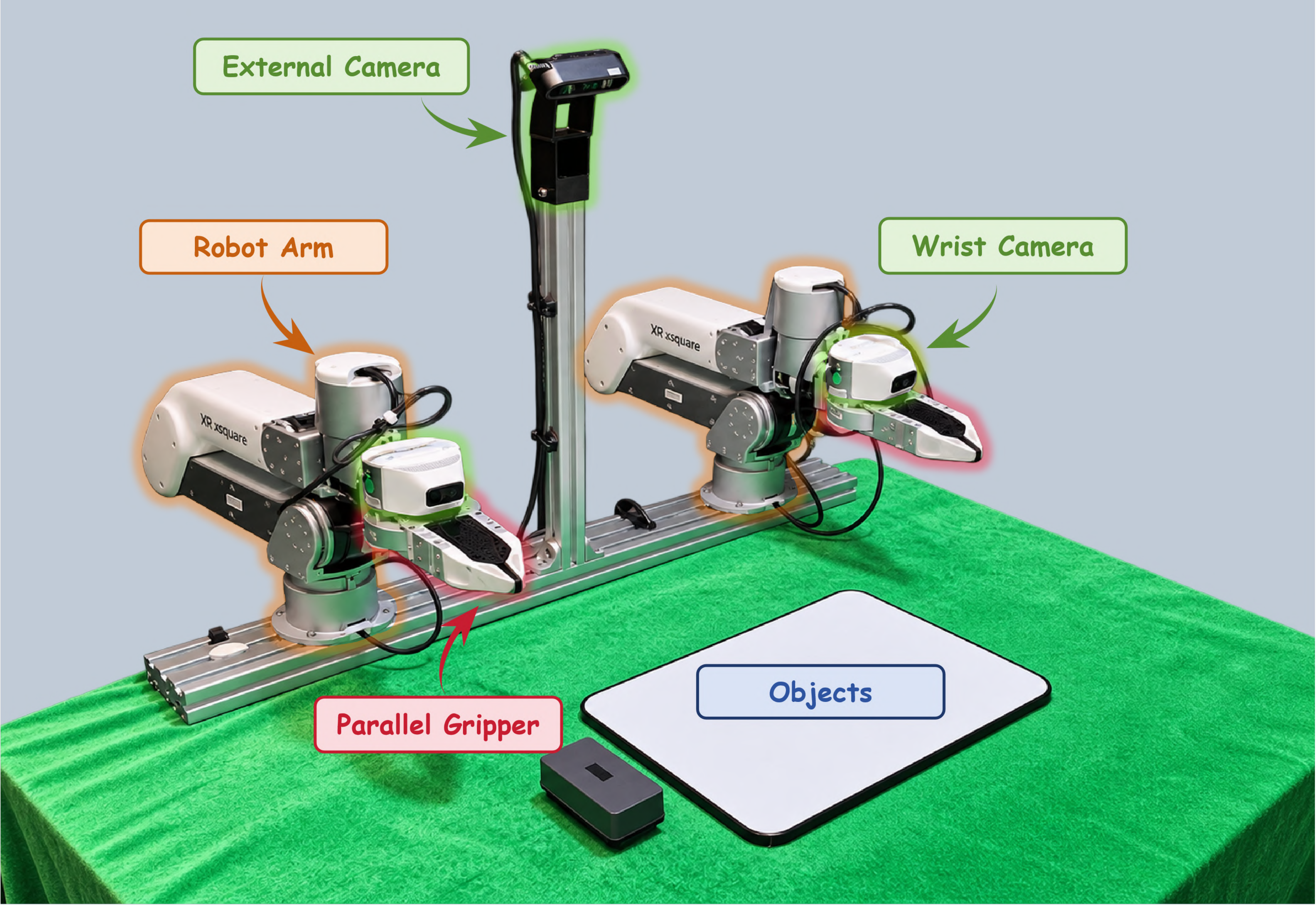}
    \end{minipage}
    \caption{Real-robot platforms. Left: single-arm Franka. Right: bimanual X Square Robot.}
    \label{fig:robot_setup}
\end{figure}

\begin{figure}[t]
    \centering
    \includegraphics[width=\columnwidth]{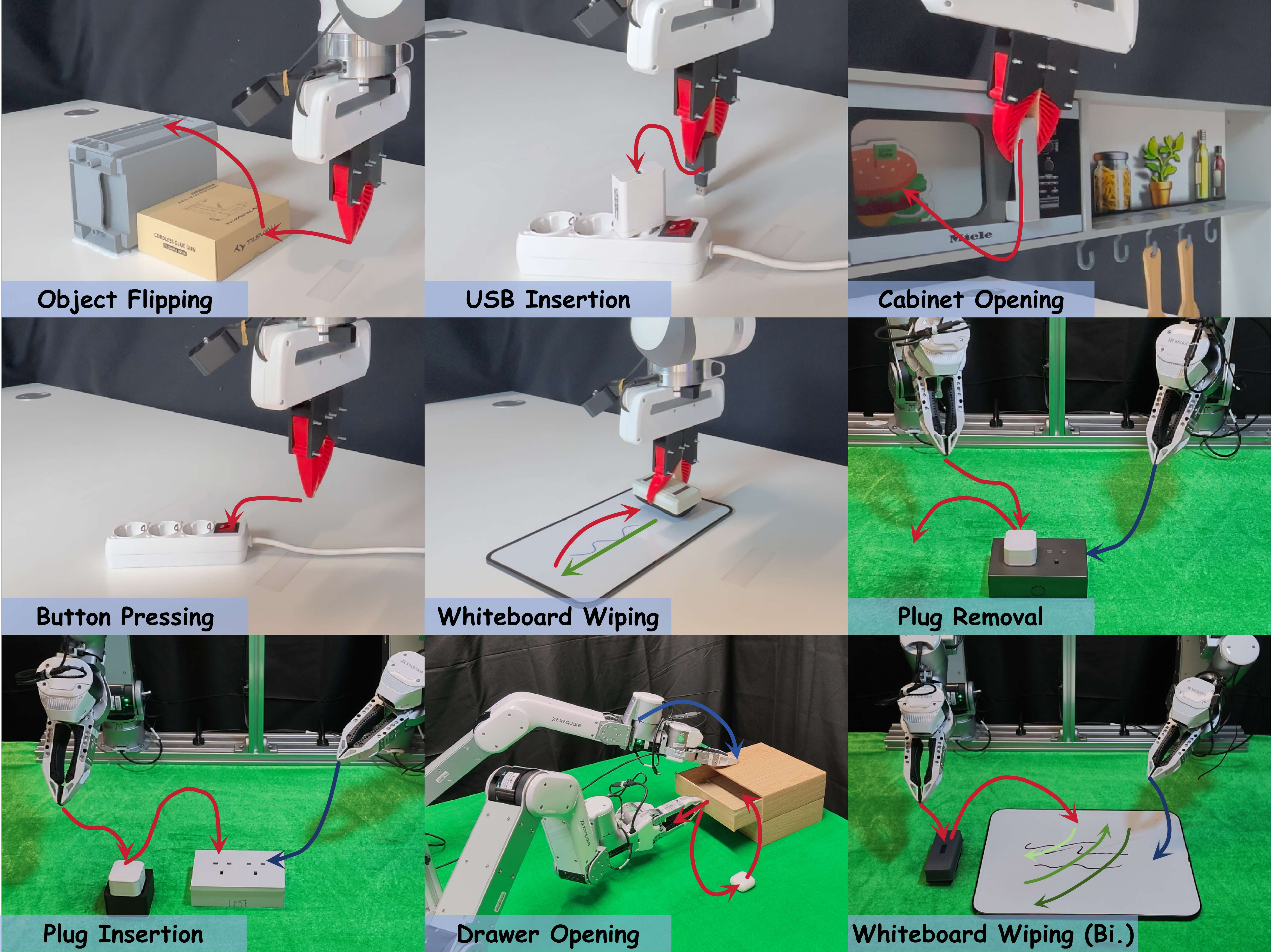}
    \caption{Nine real-robot evaluation tasks, read left to right and top to bottom. The first 5 panels show single-arm tasks; the remaining 4 show bimanual tasks.}
    \label{fig:task_overview}
\end{figure}

\noindent\textbf{Hardware, tasks, and data collection.}
The evaluation includes $5$ single-arm and $4$ bimanual tasks on the two platforms shown in Fig.~\ref{fig:robot_setup}; the tasks are summarized in Fig.~\ref{fig:task_overview}. The single-arm platform is a 7-DoF Franka Emika Panda with wrist-mounted and external Intel RealSense D405 cameras. Demonstrations are collected using a 3Dconnexion SpaceMouse. The bimanual platform is an X Square Robot with one wrist-mounted camera per arm and one shared external camera. Demonstrations are collected through leader--follower teleoperation. Following ForceVLA~\cite{yu2025forcevlaenhancingvlamodels}, both platforms use robot-provided joint-torque-based wrench estimates without additional wrist force/torque sensors.

For each task, we collect $200$ demonstrations. Multi-view images, robot states, and commanded actions are recorded at $30$~Hz, and timestamped wrench estimates at $100$~Hz.

\begin{table*}[!t]
    \centering
    \scriptsize
    \setlength{\tabcolsep}{3pt}
    \caption{Task success rate (\%) over 20 trials per method and task; Avg. is the mean across the nine tasks.}
    \label{tab:main_results}
    \begin{tabular}{lcccccccccc}
        \toprule
        & \multicolumn{5}{c}{\textbf{Single-arm}} & \multicolumn{4}{c}{\textbf{Bimanual}} & \\
        \cmidrule(lr){2-6}\cmidrule(lr){7-10}
        \textbf{Method}
        & \shortstack{Object\\Flipping}
        & \shortstack{USB\\Insertion}
        & \shortstack{Cabinet\\Opening}
        & \shortstack{Button\\Pressing}
        & \shortstack{Whiteboard\\Wiping}
        & \shortstack{Plug\\Removal}
        & \shortstack{Plug\\Insertion}
        & \shortstack{Drawer\\Opening}
        & \shortstack{Whiteboard\\Wiping}
        & \textbf{Avg.} \\
        \midrule
        $\pi_{0.5}$~\cite{intelligence2025pi05visionlanguageactionmodelopenworld} & 55.0 & 35.0 & 50.0 & 45.0 & 70.0 & 40.0 & 30.0 & 50.0 & 50.0 & 47.2 \\
        ForceVLA~\cite{yu2025forcevlaenhancingvlamodels} & 65.0 & 40.0 & 60.0 & 50.0 & 70.0 & 50.0 & 35.0 & 60.0 & 60.0 & 54.4 \\
        TA-VLA~\cite{zhang2025tavlaelucidatingdesignspace} & 65.0 & 35.0 & 60.0 & 60.0 & 75.0 & 55.0 & 40.0 & 60.0 & 65.0 & 57.2 \\
        ImplicitRDP~\cite{chen2026implicitrdpendtoendvisualforcediffusion} & 50.0 & 25.0 & 45.0 & 35.0 & 65.0 & 55.0 & 30.0 & 55.0 & 45.0 & 45.0 \\
        Temporal Teacher & 75.0 & 55.0 & 75.0 & 80.0 & 85.0 & 75.0 & 50.0 & 70.0 & 70.0 & 70.6 \\
        \textbf{ForceDelta-VLA (Ours)} & \textbf{90.0} & \textbf{80.0} & \textbf{80.0} & \textbf{85.0} & \textbf{90.0} & \textbf{80.0} & \textbf{70.0} & \textbf{80.0} & \textbf{85.0} & \textbf{82.2} \\
        \bottomrule
    \end{tabular}
\end{table*}

\noindent\textbf{Baselines and evaluation protocol.}
We compare with ForceVLA~\cite{yu2025forcevlaenhancingvlamodels}, a force-aware full-action policy that also provides our teacher backbone; $\pi_{0.5}$~\cite{intelligence2025pi05visionlanguageactionmodelopenworld}, a VLA without force/torque input; TA-VLA~\cite{zhang2025tavlaelucidatingdesignspace}, a torque-aware VLA; and ImplicitRDP~\cite{chen2026implicitrdpendtoendvisualforcediffusion}, a slow--fast visual--force policy. We also evaluate direct execution of the Stage-1 Temporal Teacher used by ForceDelta-VLA. All methods use the same task demonstrations, initial-state distributions, controller settings, and safety limits. We report success rate, peak contact force, and completion time.

\noindent\textbf{Implementation details.}
The teacher encodes a $100$-ms wrench history and predicts an action chunk with $H{=}50$ steps. The correction policy uses a causal force encoder and a shared attention module with separate force- and delay-correction heads to predict $K{=}5$ pose corrections. Deployment runs on a workstation with two NVIDIA RTX 4090 GPUs. The reference-action forward pass averages $189.7$~ms, corresponding to approximately $5$~Hz for serial inference (Table~\ref{tab:runtime}). Robot command transmission is limited to $100$~Hz; this control rate is distinct from the correction-query rate. Training and optimization details are provided in the supplementary material.

\subsection{Main Results}
\label{sec:exp_main}

\noindent\textbf{Quantitative results.}
Compared with the original ForceVLA baseline, the complete ForceDelta-VLA system improves mean success from $54.4\%$ to $82.2\%$ (Table~\ref{tab:main_results}). It also exceeds TA-VLA, $\pi_{0.5}$, and ImplicitRDP. Success improves over ForceVLA on all nine tasks, with the largest gains on USB Insertion, Button Pressing, and Plug Insertion.

Direct execution of the Stage-1 Temporal Teacher achieves $70.6\%$ mean success. ForceDelta-VLA improves mean success by about $11.7$ percentage points over this teacher. The largest gains are on USB Insertion and Plug Insertion, at $25$ and $20$ points, respectively. On Button Pressing, most of the improvement over ForceVLA is already obtained by the Temporal Teacher ($50\%$ to $80\%$); ForceDelta-VLA raises success to $85\%$.

\noindent\textbf{Contact force and completion time.}
Relative to ForceVLA, mean peak contact force over successful trials decreases by $4.2$~N on the single-arm platform and $4.3$~N on the bimanual platform, approximately $26\%$ on each (Table~\ref{tab:interaction}). Mean completion time decreases by $5.2$~s and $5.9$~s, respectively. Relative to the Temporal Teacher, the peak-force reductions are $3.3$~N and $2.2$~N, and completion time decreases by $3.9$~s and $3.4$~s.

\begin{table}[t]
    \centering
    \scriptsize
    \setlength{\tabcolsep}{2pt}
    \caption{Peak force and completion time, reported as mean $\pm$ standard deviation over successful trials pooled across tasks within each platform.}
    \label{tab:interaction}
    \begin{tabular}{lcccc}
        \toprule
        & \multicolumn{2}{c}{\textbf{Peak force} (N) $\downarrow$} & \multicolumn{2}{c}{\textbf{Completion time} (s) $\downarrow$} \\
        \cmidrule(lr){2-3}\cmidrule(lr){4-5}
        \textbf{Method} & Single & Biman. & Single & Biman. \\
        \midrule
        $\pi_{0.5}$ & $18.9 \pm 8.3$ & $20.4 \pm 9.1$ & $30.4 \pm 11.2$ & $33.1 \pm 12.8$ \\
        ForceVLA & $16.0 \pm 4.7$ & $16.8 \pm 6.2$ & $29.1 \pm 6.5$ & $31.7 \pm 9.4$ \\
        TA-VLA & $16.5 \pm 6.4$ & $15.9 \pm 4.8$ & $28.2 \pm 9.7$ & $30.0 \pm 7.1$ \\
        ImplicitRDP & $15.2 \pm 5.1$ & $14.8 \pm 4.9$ & $27.1 \pm 7.2$ & $27.4 \pm 7.8$ \\
        Temporal Teacher & $15.1 \pm 5.8$ & $14.7 \pm 4.3$ & $27.8 \pm 8.8$ & $29.2 \pm 6.9$ \\
        \textbf{\shortstack[l]{ForceDelta-VLA\\(Ours)}} & $\mathbf{11.8} \pm 3.4$ & $\mathbf{12.5} \pm 2.9$ & $\mathbf{23.9} \pm 5.4$ & $\mathbf{25.8} \pm 7.6$ \\
        \bottomrule
    \end{tabular}
\end{table}

\begin{figure}[t]
    \centering
    \includegraphics[width=\columnwidth]{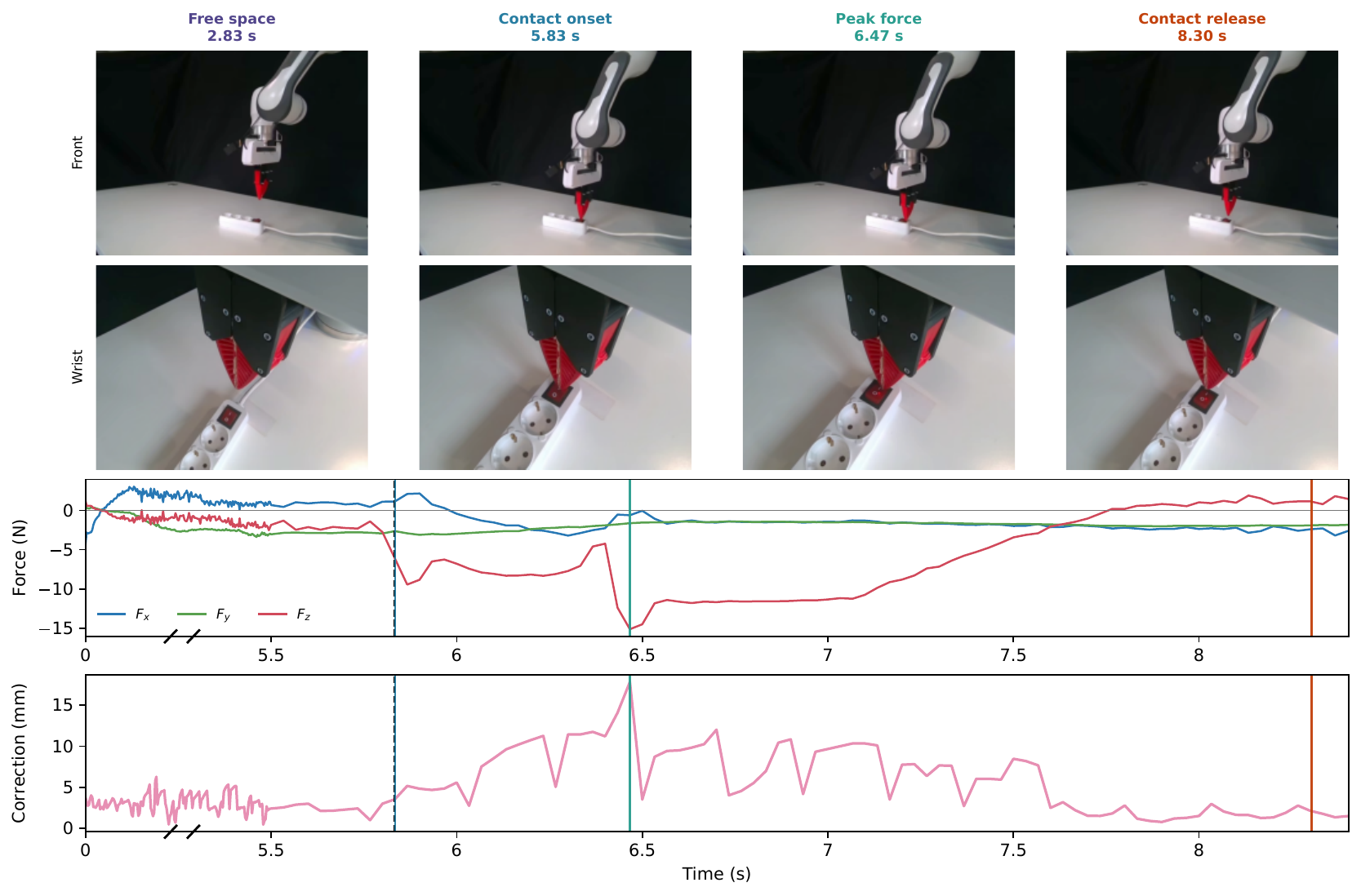}
    \caption{Real-robot execution of ForceDelta-VLA on Button Pressing. Front and wrist views show four marked moments. The curves show robot-estimated force components and translational correction magnitude. Vertical lines mark contact onset, peak force, and contact release.}
    \label{fig:qualitative_button}
\end{figure}

\noindent\textbf{Qualitative results.}
Figure~\ref{fig:qualitative_button} shows a real-robot execution of ForceDelta-VLA on Button Pressing. The correction magnitude is relatively small before contact, increases during pressing, and decreases as contact is released. This example illustrates how the correction magnitude changes across the contact phases.

\subsection{Runtime and Reference-Action Delay}
\label{sec:exp_runtime}

We measure the forward-pass latencies of the reference-action and correction pathways and evaluate task performance under additional reference-action latency.

\noindent\textbf{Forward-pass latency.}
The correction forward pass takes $2.43$~ms, below the $10$-ms interval between robot command transmissions (Table~\ref{tab:runtime}). The Temporal Teacher and reference-action forward passes both exceed that interval. By reusing a reference action, the correction pathway avoids a new generative prediction at each query.

\begin{table}[t]
    \centering
    \scriptsize
    \setlength{\tabcolsep}{2pt}
    \caption{Forward-pass latency with batch size one under the Button Pressing deployment workload, reported as mean $\pm$ standard deviation over repeated passes on the deployment workstation.}
    \label{tab:runtime}
    \begin{tabular}{lc}
        \toprule
        \textbf{Method/path} & \textbf{Forward latency (ms)} $\downarrow$ \\
        \midrule
        ImplicitRDP & $12.86 \pm 0.23$ \\
        Temporal Teacher & $176.4 \pm 5.9$ \\
        ForceDelta reference action & $189.7 \pm 6.8$ \\
        ForceDelta correction & $\mathbf{2.43} \pm 0.12$ \\
        \bottomrule
    \end{tabular}
\end{table}

\begin{figure}[t]
    \centering
    \begin{minipage}[c]{0.49\columnwidth}
        \centering
        \includegraphics[width=\linewidth]{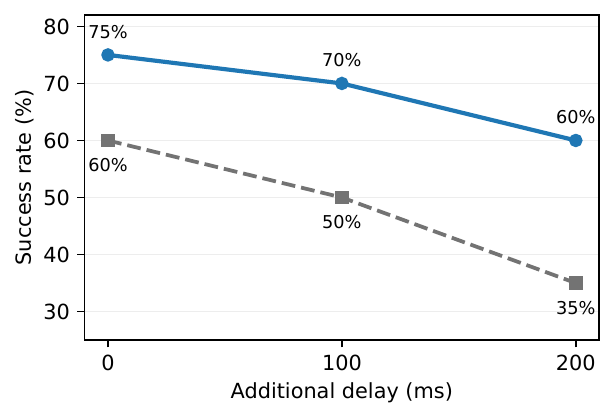}
    \end{minipage}\hfill
    \begin{minipage}[c]{0.49\columnwidth}
        \centering
        \includegraphics[width=\linewidth]{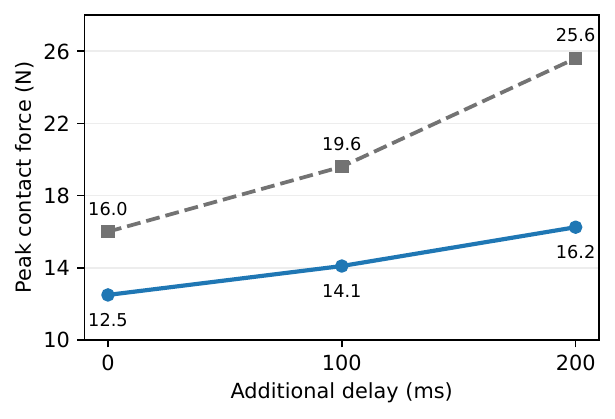}
    \end{minipage}
    \caption{Sensitivity to additional reference-action latency on USB Insertion and Plug Insertion, with 10 trials per task, method, and delay. Left: mean success rate across the two tasks. Right: mean peak force over successful trials. Blue solid lines with circular markers denote ForceDelta-VLA; gray dashed lines with square markers denote Temporal Teacher.}
    \label{fig:latency_sensitivity}
\end{figure}

\noindent\textbf{Sensitivity to reference-action delay.}
On USB Insertion and Plug Insertion, increasing additional reference-action latency from $0$ to $200$~ms reduces success by $15$ percentage points for ForceDelta-VLA and $25$ points for the Temporal Teacher (Fig.~\ref{fig:latency_sensitivity}). Mean peak force over successful trials increases by $3.7$~N and $9.6$~N, respectively. These results support updating corrections between reference-action updates.

\subsection{Ablation Studies}
\label{sec:exp_ablation}

The ablations ask three questions: how the correction target should be defined, whether the fast policy should predict separate corrections, a combined correction, or complete actions, and how training should account for asynchronous execution (Table~\ref{tab:ablation}). All variants are evaluated on USB Insertion, single-arm Whiteboard Wiping, Plug Insertion, and Drawer Opening.

\noindent\textbf{Ablation setup.}
Execution ablations retain the trained model: \textbf{Reference action only} disables both corrections; \textbf{w/o force correction} disables the force output; \textbf{w/o learned delay correction} disables the full delay output, including reference-state alignment; and \textbf{Zeroed correction input} supplies an all-zero force history while retaining the force branch. The remaining variants retrain the affected policy. Their changes are described with the corresponding results below.

\begin{table}[t]
    \centering
    \scriptsize
    \setlength{\tabcolsep}{2pt}
    \caption{Ablations on four tasks, with 20 trials per variant and task. Results are grouped by platform. Peak force: mean $\pm$ standard deviation over successful trials.}
    \label{tab:ablation}
    \begin{tabular}{lcccc}
        \toprule
        & \multicolumn{2}{c}{\textbf{Success rate} (\%) $\uparrow$}
        & \multicolumn{2}{c}{\textbf{Peak force} (N) $\downarrow$} \\
        \cmidrule(lr){2-3}\cmidrule(lr){4-5}
        \textbf{Variant} & \textbf{Single} & \textbf{Biman.}
        & \textbf{Single} & \textbf{Biman.} \\
        \midrule
        \multicolumn{5}{l}{\emph{Execution/component ablations}} \\
        Reference action only & $62.5$ & $50.0$ & $16.7 \pm 6.9$ & $18.2 \pm 7.6$ \\
        w/o force correction & $70.0$ & $57.5$ & $15.4 \pm 5.8$ & $17.1 \pm 6.3$ \\
        \shortstack[l]{w/o learned\\delay correction} & $77.5$ & $67.5$ & $13.2 \pm 4.1$ & $14.5 \pm 5.0$ \\
        Zeroed correction input & $65.0$ & $55.0$ & $15.9 \pm 6.4$ & $17.7 \pm 7.0$ \\
        \textbf{ForceDelta-VLA (full)} & $\mathbf{85.0}$ & $\mathbf{75.0}$ & $\mathbf{11.7} \pm 3.1$ & $\mathbf{12.8} \pm 3.8$ \\
        \midrule
        \multicolumn{5}{l}{\emph{Supervision/decomposition ablations}} \\
        \shortstack[l]{Demonstration-derived\\correction target} & $72.5$ & $57.5$ & $14.6 \pm 5.2$ & $16.5 \pm 6.1$ \\
        \shortstack[l]{Zero-wrench\\teacher query} & $65.0$ & $55.0$ & $16.1 \pm 6.7$ & $17.3 \pm 6.5$ \\
        w/o schedule replay & $77.5$ & $62.5$ & $13.8 \pm 4.7$ & $15.6 \pm 5.8$ \\
        \shortstack[l]{Single combined-\\correction head} & $75.0$ & $65.0$ & $13.5 \pm 4.3$ & $14.9 \pm 4.9$ \\
        \shortstack[l]{Fast full-action\\distillation} & $67.5$ & $52.5$ & $15.2 \pm 5.6$ & $17.9 \pm 7.4$ \\
        \textbf{ForceDelta-VLA (full)} & $\mathbf{85.0}$ & $\mathbf{75.0}$ & $\mathbf{11.7} \pm 3.1$ & $\mathbf{12.8} \pm 3.8$ \\
        \bottomrule
    \end{tabular}
\end{table}

\begin{samepage}
\noindent\textbf{How should correction supervision be defined?}
\textbf{Demonstration-derived correction target} replaces the force target with the demonstrated action minus the force-agnostic teacher prediction at the same query state. Success decreases by $12.5$ and $17.5$ percentage points on the single-arm and bimanual platforms, respectively. This comparison favors defining force supervision through teacher predictions matched in context, state, and sampling noise. \textbf{Zero-wrench teacher query}, used for both reference generation and target construction, also lowers success, supporting the learned force-agnostic mode.

\end{samepage}

\noindent\textbf{What should the fast policy predict?}
Removing the force correction reduces success by $15$ and $17.5$ percentage points and increases mean peak force over successful trials by $3.7$ and $4.3$~N on the single-arm and bimanual platforms, respectively. The force branch thus contributes to both task completion and contact-force reduction. Removing both corrections lowers success further. Zeroing the force history also performs worse than disabling the force output, indicating that the active branch depends on force feedback.

\textbf{Single combined-correction head} trains one head on the summed target and reduces success by $10$ percentage points on both platforms, favoring separate supervision and prediction of the two corrections. \textbf{Fast full-action distillation} of the teacher's pose actions also performs worse under the same compact student architecture, inputs, and update schedule. This favors reusing reference actions and assigning the student to predict their corrections.

\noindent\textbf{How does asynchronous adaptation contribute?}
Disabling the learned delay correction reduces success by $7.5$ percentage points on both platforms and increases peak force, supporting the adjustment for reference-action mismatch and reference-state alignment. Omitting \textbf{schedule replay} reduces success by $7.5$ and $12.5$ percentage points on the single-arm and bimanual platforms, respectively. The student therefore also benefits from training on delayed references.

\subsection{Generalization to Unseen Objects}
\label{sec:exp_physical_generalization}

We evaluate USB Insertion, Object Flipping, and bimanual Whiteboard Wiping with the unseen objects shown in Fig.~\ref{fig:physical_generalization_objects}.

\begin{table}[!htbp]
    \centering
    \includegraphics[width=0.85\columnwidth]{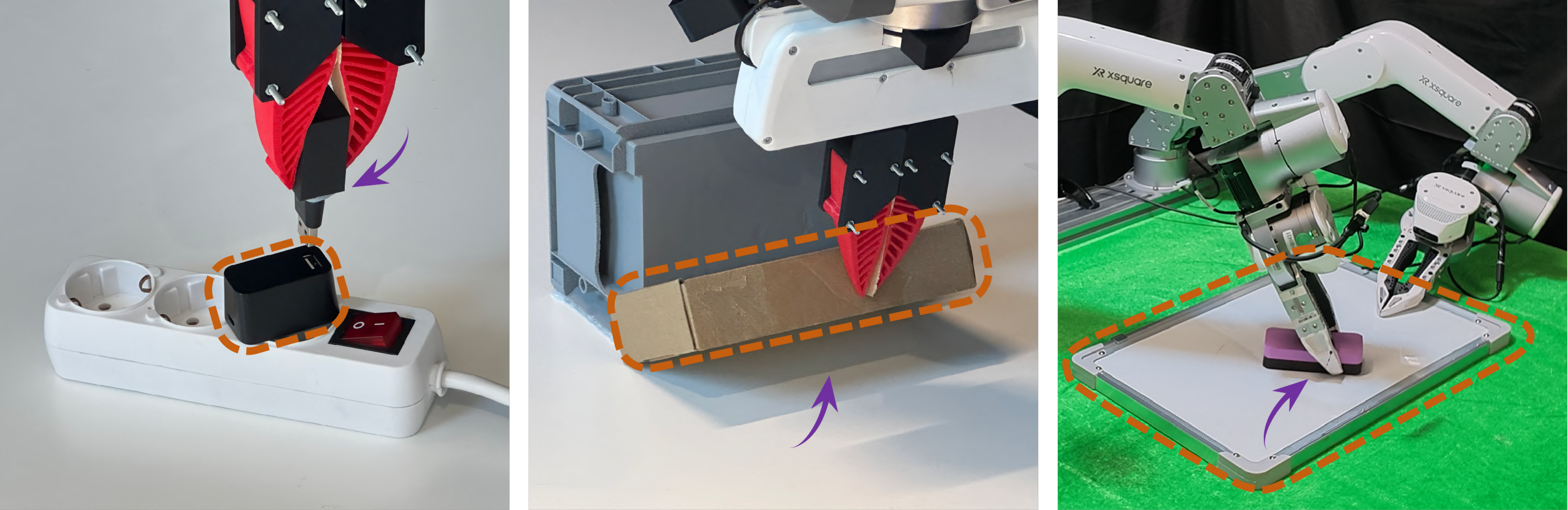}
    \begingroup
    \captionof{figure}{Unseen objects used for generalization evaluation. From left to right: USB Insertion, Object Flipping, and bimanual Whiteboard Wiping.}
    \label{fig:physical_generalization_objects}
    \endgroup

    \scriptsize
    \setlength{\tabcolsep}{5pt}
    \caption{Generalization with unseen objects, with 10 trials per method and task. Success rate is computed across the three tasks. Peak force: mean $\pm$ standard deviation over successful trials.}
    \label{tab:physical_generalization}
    \begin{tabular}{lcc}
        \toprule
        \textbf{Method} & \textbf{Success rate} (\%) $\uparrow$ & \textbf{Peak force} (N) $\downarrow$ \\
        \midrule
        ImplicitRDP & $16.7$ & $45.9 \pm 10.5$ \\
        Temporal Teacher & $40.0$ & $21.8 \pm 5.2$ \\
        \textbf{ForceDelta-VLA (Ours)} & $66.7$ & $13.6 \pm 3.7$ \\
        \bottomrule
    \end{tabular}
\end{table}

On unseen objects, ForceDelta-VLA improves success over the Temporal Teacher by $26.7$ percentage points and reduces mean peak force over successful trials by $8.2$~N (Table~\ref{tab:physical_generalization}). The new USB port is particularly challenging for the Temporal Teacher. For ImplicitRDP, failures often occur while establishing contact: the policy misses the replacement eraser during grasping or fails to initiate the downward insertion motion.

\subsection{Failure Analysis}
\label{sec:exp_limitations}

In failed USB and plug insertion trials, the robot often reached the target region but contacted the rim or entered with a small orientation error. Once the connector jammed, the reference action often continued advancing, and the corrected motion did not retreat far enough to clear the jam and retry.

In cabinet and drawer opening, pulling sometimes began before stable contact with the handle was established. Recovery required releasing contact, repositioning the gripper, or approaching again. Wiping and object-flipping failures involved contact on an unintended side or loss of the intended contact geometry, requiring a different approach direction.

Reference-update delay further reduced insertion success (Fig.~\ref{fig:latency_sensitivity}). During this delay, the robot could enter a new contact state while the student continued correcting a chunk generated before that change. These failures point to the need for reference updates that guide the robot to move back, restore contact, or change the approach direction.

\FloatBarrier

\section{Conclusion and Future Work}
\label{sec:conclusion}
We presented ForceDelta-VLA, a correction-distillation framework for contact-rich manipulation. Paired force-conditioned and force-agnostic teacher predictions define force-correction targets, while delay-correction targets account for reference-action mismatch and reference-state alignment. The correction policy is trained with asynchronous schedule replay and updates pose corrections between reference-action updates. Across nine real-robot tasks, ForceDelta-VLA achieved $82.2\%$ mean success, compared with $54.4\%$ for ForceVLA and $70.6\%$ for direct execution of the Stage-1 Temporal Teacher. Mean peak contact force over successful trials was approximately $26\%$ lower than that of ForceVLA on both platforms.

Future work will incorporate high-precision force/torque sensors for fine manipulation tasks with force feedback.


{
\small
\bibliographystyle{IEEEtran}
\bibliography{main}
}

\end{document}